\documentclass[sigconf]{acmart}

\usepackage{booktabs} 
\usepackage{url}
\usepackage{algorithm}
\usepackage[noend]{algpseudocode}
\usepackage{setspace}
\usepackage{multirow}
\usepackage{bbm}

\usepackage{xspace}
\usepackage{balance}

\usepackage[font={footnotesize}]{caption}

\newcommand{\inspect}{\mathbf{h}}

\newcommand{\replace}{\mathbf{r}}

\newcommand{\pool}{\mathcal{P}}

\def\statmodel{\textsc{StatisticalModel}\xspace}
\def\druleHVI{\textsc{InspectionDecisionRule}\xspace}
\def\druleSLR{\textsc{ReplacementDecisionRule}\xspace}
\def\activerem{\textsc{ActiveRemediation}\xspace}
\makeatletter
\def\BState{\State\hskip-\ALG@thistlm}
\makeatother

\newcommand{\jake}[1]{\textcolor{blue}{\textbf{(jake)} #1}}

\newcommand{\urlformat}[1]{\emph{ #1}}

\setcopyright{rightsretained}
\acmDOI{10.475/123_4}

\acmISBN{123-4567-24-567/08/06}

\copyrightyear{2018}
\acmYear{2018}
\setcopyright{acmcopyright}
\acmConference{KDD '18}{}{August 19-23, 2017, London, UK}

\acmArticle{4}
\acmPrice{15.00}

\author{Jacob Abernethy}
\affiliation{%
  \institution{Georgia Institute of Technology \&}
  \institution{University of Michigan}
}
\email{prof@gatech.edu}

\author{Alex Chojnacki}
\affiliation{%
  \institution{University of Michigan}
}
\email{thealex@umich.edu}

\author{Arya Farahi}
\affiliation{%
  \institution{University of Michigan}
}
\email{aryaf@umich.edu}

\author{Eric Schwartz}
\affiliation{%
  \institution{University of Michigan}
}
\email{ericmsch@umich.edu}

\author{Jared Webb}
\affiliation{%
  \institution{Brigham Young University}
}
\email{webb@mathematics.byu.edu}

\begin{document}
\title{ActiveRemediation: The Search for Lead Pipes in Flint, Michigan}

\copyrightyear{2018}
\acmYear{2018}
\setcopyright{acmcopyright}
\acmConference[KDD '18]{The 24th ACM SIGKDD International Conference
on Knowledge Discovery \& Data Mining}{August 19--23, 2018}{London,
United Kingdom}
\acmBooktitle{KDD '18: The 24th ACM SIGKDD International Conference on
Knowledge Discovery \& Data Mining, August 19--23, 2018, London, United
Kingdom}
\acmPrice{15.00}
\acmDOI{10.1145/3219819.3219896}
\acmISBN{978-1-4503-5552-0/18/08}

\renewcommand{\shortauthors}{J. Abernethy et al.}

\begin{abstract}
  We detail our ongoing work in Flint, Michigan to detect pipes made of lead and other hazardous metals. After elevated levels of lead were detected in residents' drinking water, followed by an increase in blood lead levels in area children, the state and federal governments directed over \$125 million to replace water service lines, the pipes connecting each home to the water system. In the absence of accurate records, and with the high cost of determining buried pipe materials, we put forth a number of predictive and procedural tools to aid in the search and removal of lead infrastructure. Alongside these statistical and machine learning approaches, we describe our interactions with government officials in recommending homes for both inspection and replacement, with a focus on the statistical model that adapts to incoming information. Finally, in light of discussions about increased spending on infrastructure development by the federal government, we explore how our approach generalizes beyond Flint to other municipalities nationwide.
\end{abstract}

%
%

\ccsdesc[500]{Information systems~Data analytics}
\ccsdesc[300]{Machine learning~Applied computing}

\keywords{Water Infrastructure; Flint Water Crisis; Risk Assessment; Machine Learning; Active Learning; Public Policy}

\maketitle

\newpage

\linespread{1.5}

\section{Introduction}

The story of the Flint Water Crisis is long and has many facets, involving government failures, public health challenges, and social and economic justice. As Flint struggled financially after the 2008 housing crisis, the state of Michigan installed emergency managers to implement several cost saving measures. One of these actions was to switch Flint's drinking water source from the Detroit system to the local Flint river in April 2014. The new water had different chemical characteristics which were overlooked by water officials. Of course many water systems have lead pipes, but these pipes are typically coated with layers of deposits, and the water is treated appropriately in order to prevent corrosion and the leaching of heavy metals.
City officials failed to follow such necessary procedures, the pipes began to corrode, Flint's drinking water started to give off a different color and smell \cite{Mlive-toxicleadgetsintoFlintwater:url}, and Flint residents were exposed to elevated levels of lead for nearly two years before the problems received proper attention. In August 2015 environmental engineers raised alarm bells about contaminated water\footnote{Prior work by the authors involved estimation of water lead contamination \cite{abernethy2016flint}.} \cite{torrice2016lead}, not long after a pediatrician observed a jump in the number of Flint children with high blood lead levels\footnote{For further analysis of blood lead levels, see \cite{potash2015predictive}}\cite{hanna2016elevated}, and by January 2016 the Flint Water Crisis was international news.

As attention to the problem was growing, government officials at all levels got involved in managing the damage and pushing recovery efforts.  In looking for the primary source of lead in Flint's water distribution, attention turned to Flint's \emph{water service lines}, the pipes that connect homes to the city water system. These service lines are hypothesized to be the prime contributor to lead water contamination across the United States \citep{sandvig2008contribution}. Service lines, therefore, became a top priority for the City of Flint in February 2016. The Michigan state legislature eventually appropriated \$27M towards the expensive process of replacing these lines at large scale; later the U.S. Congress allocated another nearly \$100M towards the recovery effort. The group directed to execute the replacement program was called Flint Fast Action and Sustainability program (FAST Start), and their task was to remove as many hazardous service lines as possible up to funding levels.

The primary obstacle that the FAST Start team has faced throughout their work is uncertainty about the locations of lead or galvanized pipes.  Although the U.S. Environmental Protection Agency requires cities to maintain an active inventory of lead service line locations, Flint failed to do so. Service line materials are in theory documented during original construction or renovation, but in practice these records are often incomplete or lost. Most importantly, because the information is buried underground, it is costly to determine the material composition of even a single pipe. Digging up an entire water service line pipe under a resident's yard costs thousands of dollars.
City officials were uncertain about the total number of hazardous service lines in the city, with estimates ranging from a few thousand to tens of thousands.
Uncertainty about the service line material for individual homes has dramatic cost implications, as construction crews will end up excavating pipes that do not need to be replaced. These questions---how many pipes need to be replaced and which home's pipes need remediation---are at the core of the work in this paper.

\begin{figure}[!ht]
\begin{center}
\includegraphics[width=1.1in]{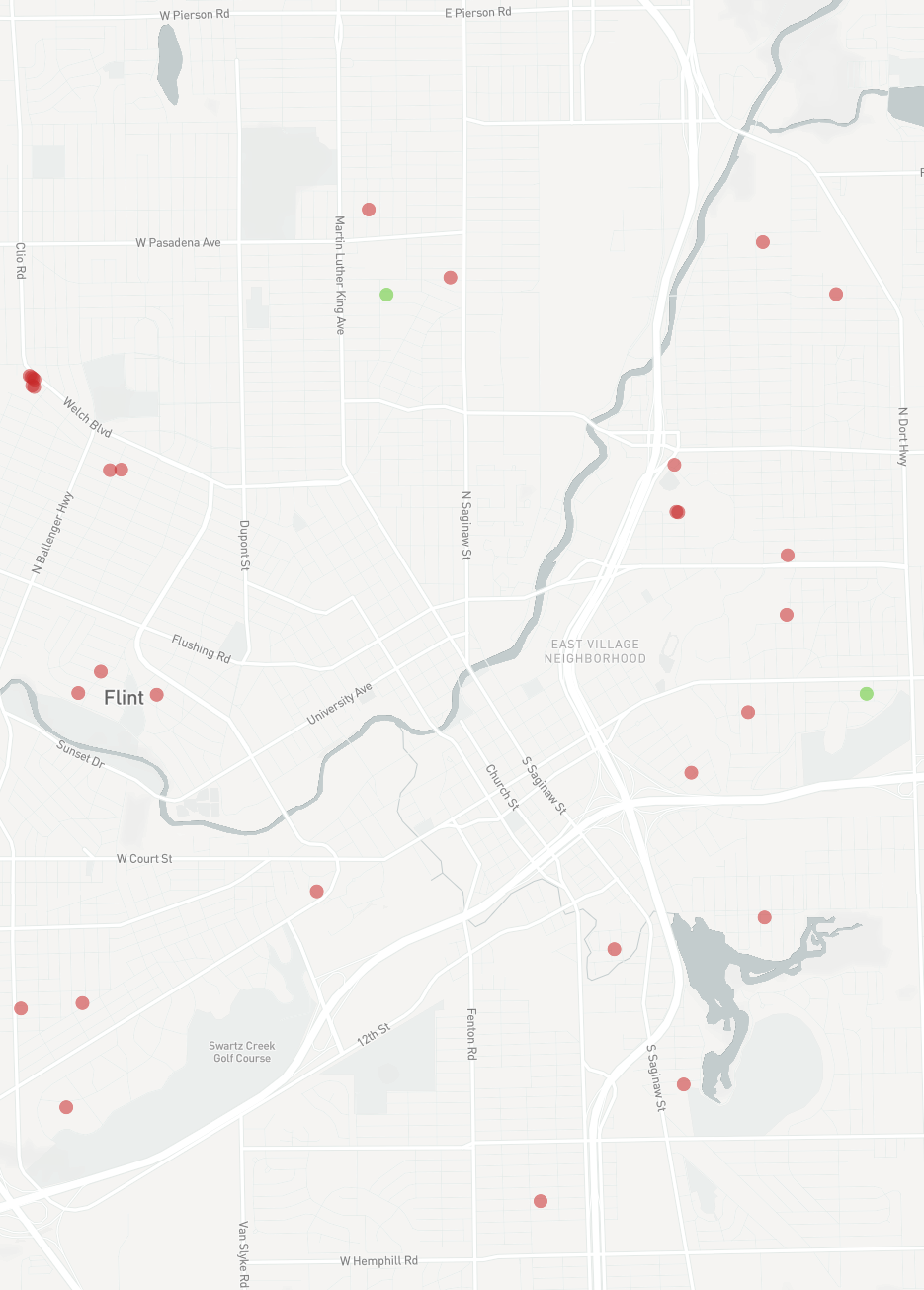}
\includegraphics[width=1.1in]{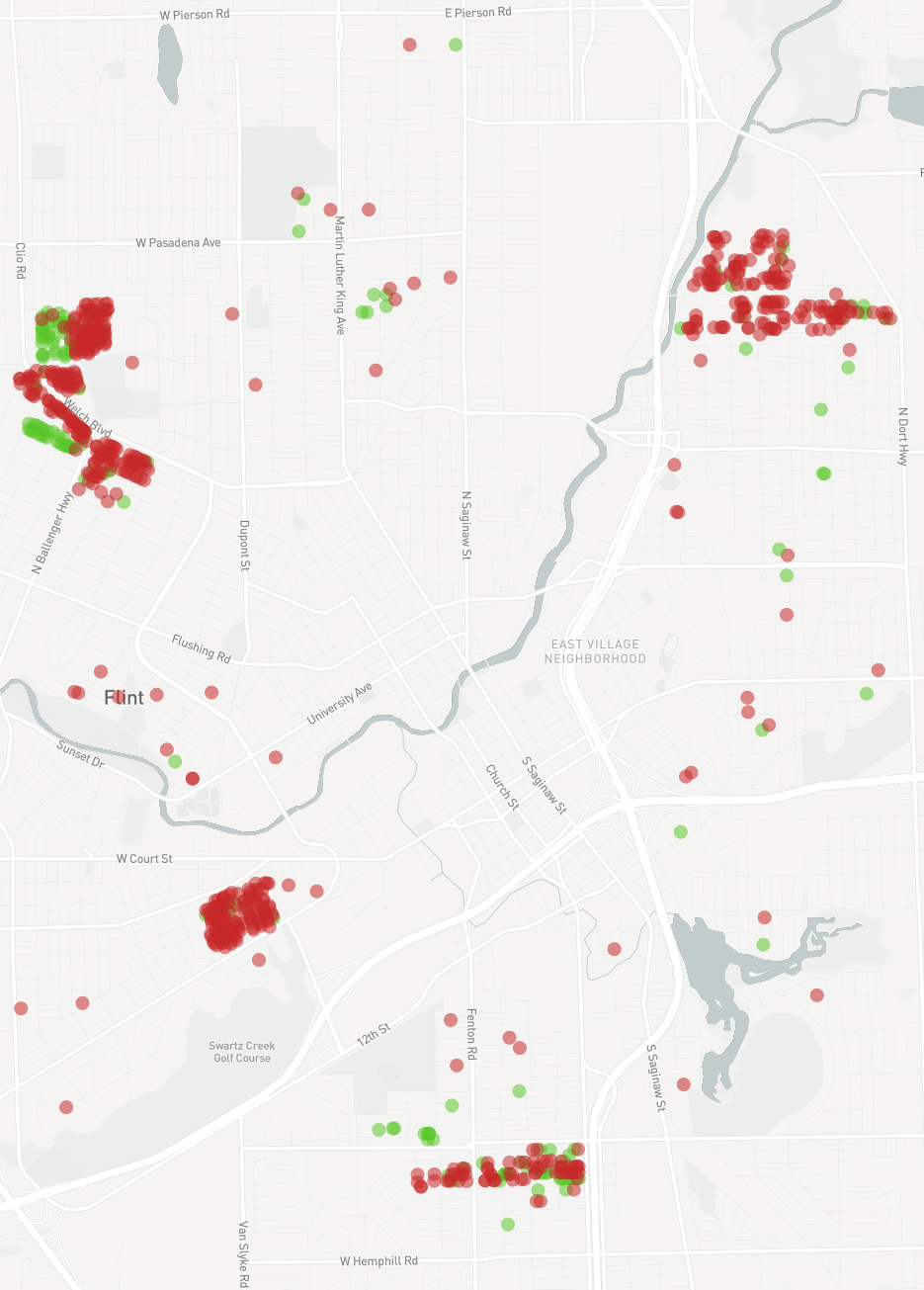}
\includegraphics[width=2.15in]{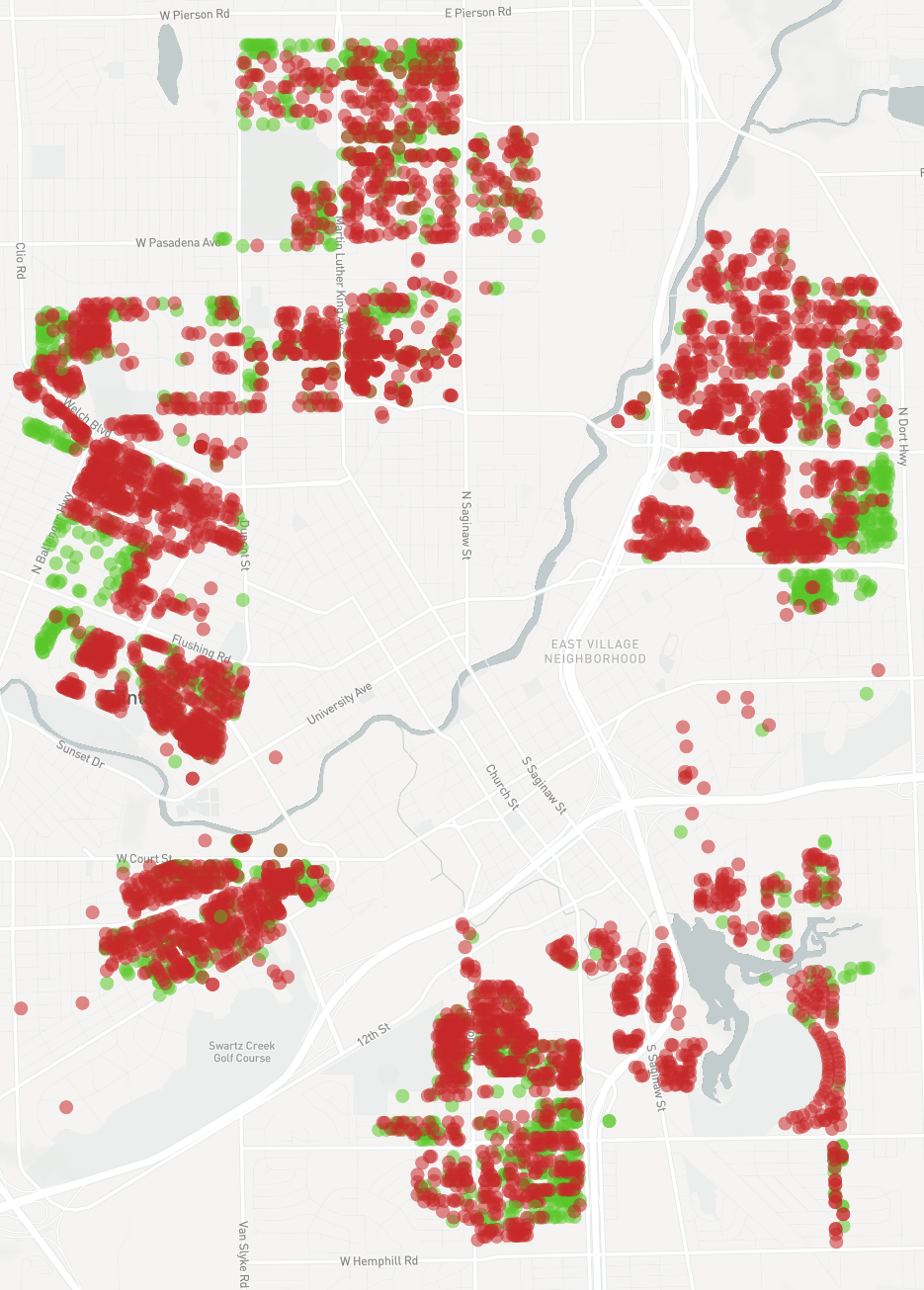}
\caption{
  Progress of the replacement program. By March of 2016, only 36 homes had undergone replacement (top left); by December 2016, a total of 762 homes either been inspected or fully replaced (top right); as of September of 2017, this had grown to a total of 6,506 homes (bottom). Homes labeled green were selected for replacement but were deemed safe after copper lines were discovered by contractors.} \label{fig:maps_by_time}
\end{center}
\end{figure}

Beginning in 2016, our team began collaborating directly with Flint city officials, analyzing the available data to provide statistical and algorithmic support to guide decision making and data collection, focusing primarily on the work of the FAST Start pipe replacement efforts. By assembling a rich suite of datasets, including thousands of water samples, information on pipe materials, and city records, we have been able to accurately estimate the locations of homes needing service line replacement, as well as those with safe pipes, in order to target recovery resources more effectively. Specifically, we have combined statistical models with active learning methods that sequentially seek out homes with hazardous water infrastructure. Along the way we have developed web-based and mobile applications for coordination among government offices, contractors, and residents. Over time, the number of homes' service lines inspected and replaced has increased, as seen in Figure \ref{fig:maps_by_time}.

In the present paper, we detail the challenges faced by decision-makers in Flint, and describe our nearly two years of work to support their efforts. With the understanding that many municipalities across the US and the world will need to undertake similar steps, we propose a generic framework which we call \activerem, that lays out a data driven approach to efficiently replace hazardous water infrastructure at large scale. We describe our implementation of \activerem in Flint, and describe the empirical performance and potential for cost savings. To our knowledge, this is the first attempt to predict the pipe materials house-by-house throughout a water system using incomplete data and also the first to propose a statistical method for adaptively selecting homes for inspection to replace hazardous materials in the most cost effective manner. This work illustrates a holistic, data-driven approach which can be replicated in other cities, thereby enhancing water infrastructure renovation effort with data-driven approaches.

\emph{Key Results.} Among our main results, we emphasize that our predictive model is empirically accurate for estimating whether a Flint home's pipes are safe/unsafe, with an AUROC score of nearly 0.92, and a true positive rate of 97\%. Since our approach involves a sequential protocol that manages the selection of homes for inspection and replacement based on our statistical model, we are also able to compare the model's total remediation cost to that of the existing protocol of officials. \activerem reduces the costly error rate (fraction of unnecessary replacements) to 2\%, lowering the effective cost of each replacement by 10\% and yielding about \$10M in potential savings.

 \emph{Methodology.} Let us now give a birds-eye view of our methodological template. \activerem manages the inspection and replacement of water service lines across a city, with the long-term objective of replacing the largest number of hazardous pipes in a city under a limited budget. The formal in-depth exposition of this framework will be given in Section~\ref{sec:overall_framework}.

\begin{algorithm}
\caption{\activerem}\label{alg:overview}
\begin{algorithmic}[1]
\State Input: parcel data, available labeled homes
\For{decision period $t = 1, \ldots, T$:}
    \State Predict hazardous/safe material via \statmodel
    \If{Budget remaining} querying \statmodel,
        \State Generate inspections via \druleHVI\
        \State Generate replacements via \druleSLR\
        \State Input observed data to \statmodel
    \EndIf
\EndFor
\end{algorithmic}
\end{algorithm}

Since the process of identifying and replacing these lines around a city is naturally sequential, the decisions and observations made earlier in the process ought to guide decisions made at future stages. With this in mind, our framework continuously maintains three subroutines that are updated as data arrives. Following the outline in Algorithm~\ref{alg:overview}, the first of these is a \statmodel, that generates probabilistic estimates of the material type of both the public and private portion of each home's service lines. The input of this model is property data, water test results, historical records, and observed service line materials. The second subroutine is \druleHVI, the decision procedure that that generates a (randomized) set of homes for inspection. This should be viewed as an \emph{active learning} protocol, with the goal of ``focused exploration.'' The third routine, \druleSLR, makes decisions as to which homes should receive line replacements; for reasons we discuss below, we typically assume that \druleSLR is a \emph{greedy} algorithm.

\emph{Roadmap.} This paper is structured as follows. We begin in Section~\ref{sec:data_availability} by laying out the datasets available to us, with the story given chronologically to describe the shifting narrative as information emerged. We then explain the \activerem framework in greater detail in Section~\ref{sec:overall_framework}, and sketch out the statistical model mixed with the prediction, inspection, and decision-making framework. In Section~\ref{sec:empirical_flint} we employ \activerem on the data available in Flint, to show the empirical performance of our proposed methods in an actual environment, as well as in a simulated environment leveraged from Flint's data. We finish by detailing the potential for significant cost savings using our approach.

\section{Emerging Data Story of Flint's Pipes}
\label{sec:data_availability}

We now describe the various sources of data and the timeline during which these became available. This is summarized in Table \ref{tbl:data_history} and more precise chronology is given throughout this section. More details will be available in the full version of this work.

\subsection{Pre-crisis Information -- Through mid-2015} \label{sec:precrisisdata}

In this section, we explain the relevant datasets that had been collected and maintained prior to the water crisis. This information, as we discovered later, was limited in both depth and quality.

\begin{table}[!t]
\footnotesize{
    \begin{tabular}{ p{0.45in}p{2.7in} }
    \hline
    Date & Description \\
    \hline
    2016 Feb. & Attributes for all 55k parcels provided by the City of Flint \\
    2016 Feb. & SL records digitized by M. Kaufman at UM Flint GIS \\
    2016 March & Pilot Program, 36 homes visited, 33 SLs replaced\\
    2016 June & Michigan DEQ provides SL private-portion inspections dataset\\
    2016 Sept. & Phase One begins, contractors use our mobile data collection app \\
    2016 Oct. & Fast Start begins hydrovac inspections to verify some home SLs\\
    2016 Oct. & Congress appropriated \$100M in WIIN Act. \\
    2016 Dec. & Fast Start \& authors release report: 20-30k replacements needed\\
    2017 March & Federal court orders 18k homes to receive SL replacement by 2019 \\
    2017 Sept. & Fast Start replaced 4,419 hazardous service lines so far, identifying  composition of a total of 6,506 homes.\\
    \hline
    \end{tabular}
} 
\caption{Timeline of service line data availability}\label{tbl:data_history}
\end{table}

\subsubsection{Parcel Data}

\begin{figure}[h]
\begin{center}
\includegraphics[width=1.6in]{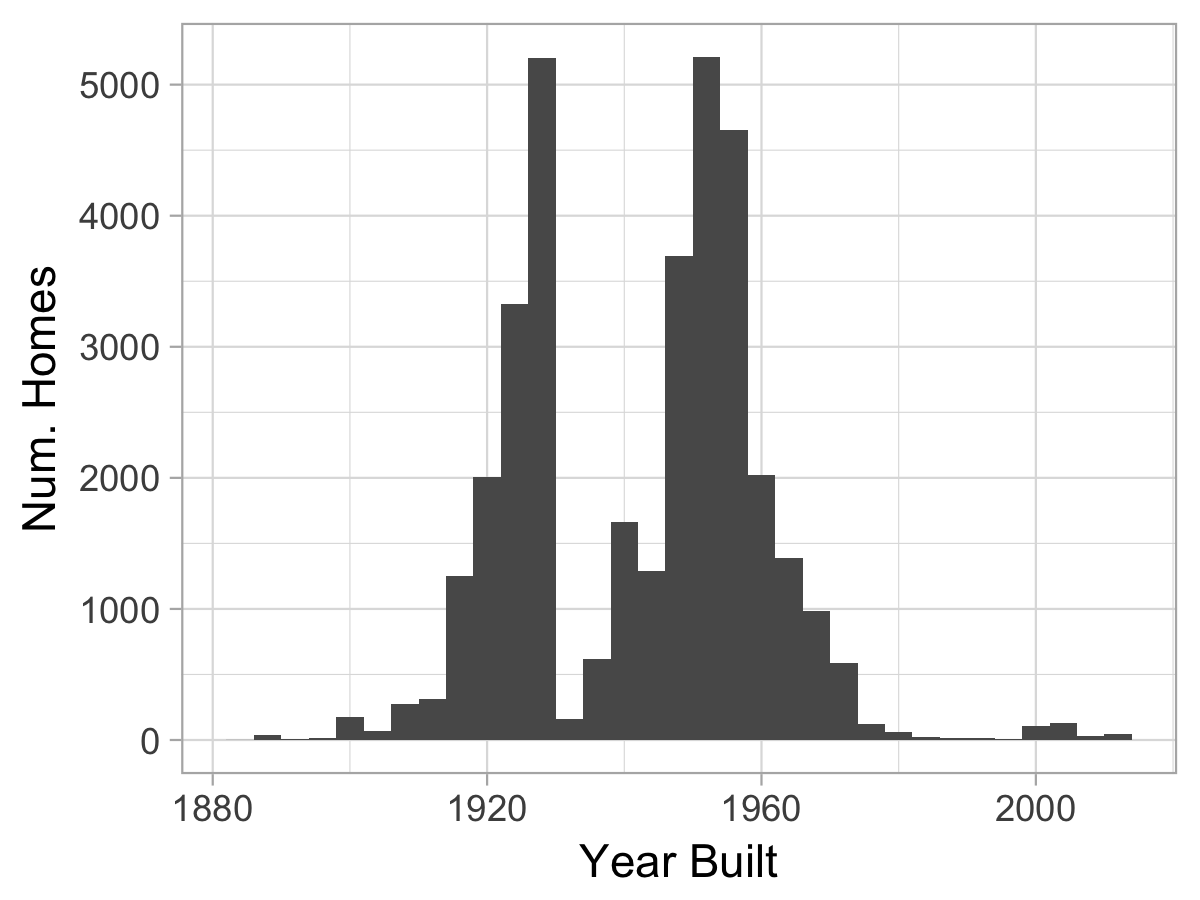}
\includegraphics[width=1.6in]{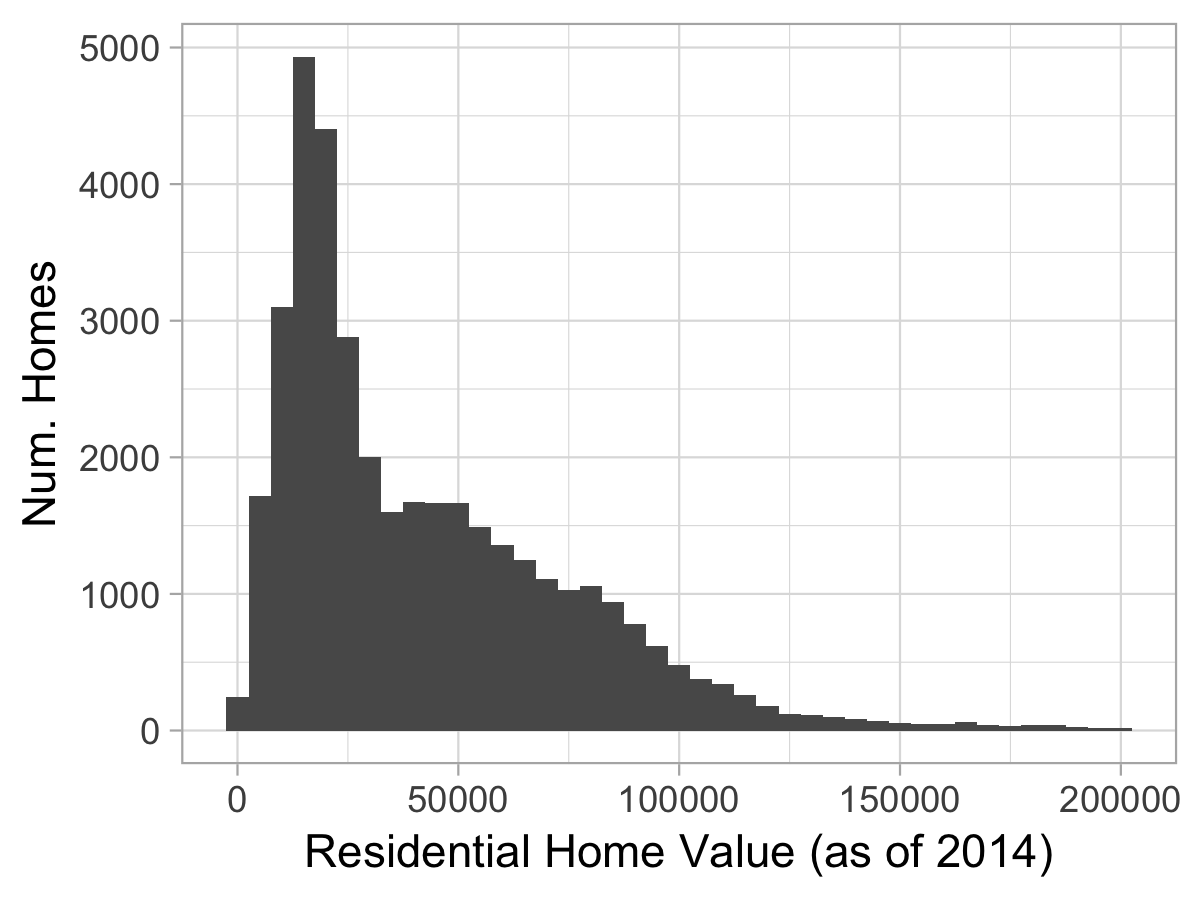}
\caption{From city parcel data, distribution of home construction by year (left) and building value by dollar (right).  The majority of the housing stock in Flint was built when it was a major automobile manufacturing hub, before current regulations about lead infrastructure were in place.  Flint has experienced significant economic decline in recent years, leading to depressed real estate prices.}
\label{fig:parcel_summaries}
\end{center}
\end{figure}

The city of Flint generously provided us with a dataset describing each of the 55,893 parcels in the city. These data include a unique identifier for each parcel and a set of columns describing City-recorded attributes of each home, such as the property owner, address, value, and building characteristics. A complete list of the parcel features is discussed in our previous work \cite{chojnackietal2017kdd}. The distributions of the age of homes and their estimated values (Figure \ref{fig:parcel_summaries}) tell an important story about the kinds of properties in Flint. 

\subsubsection{City Records of Service Lines}
\label{sec:cityrecords}

\begin{figure}[!ht]
\begin{center}
\includegraphics[height=2.5in]{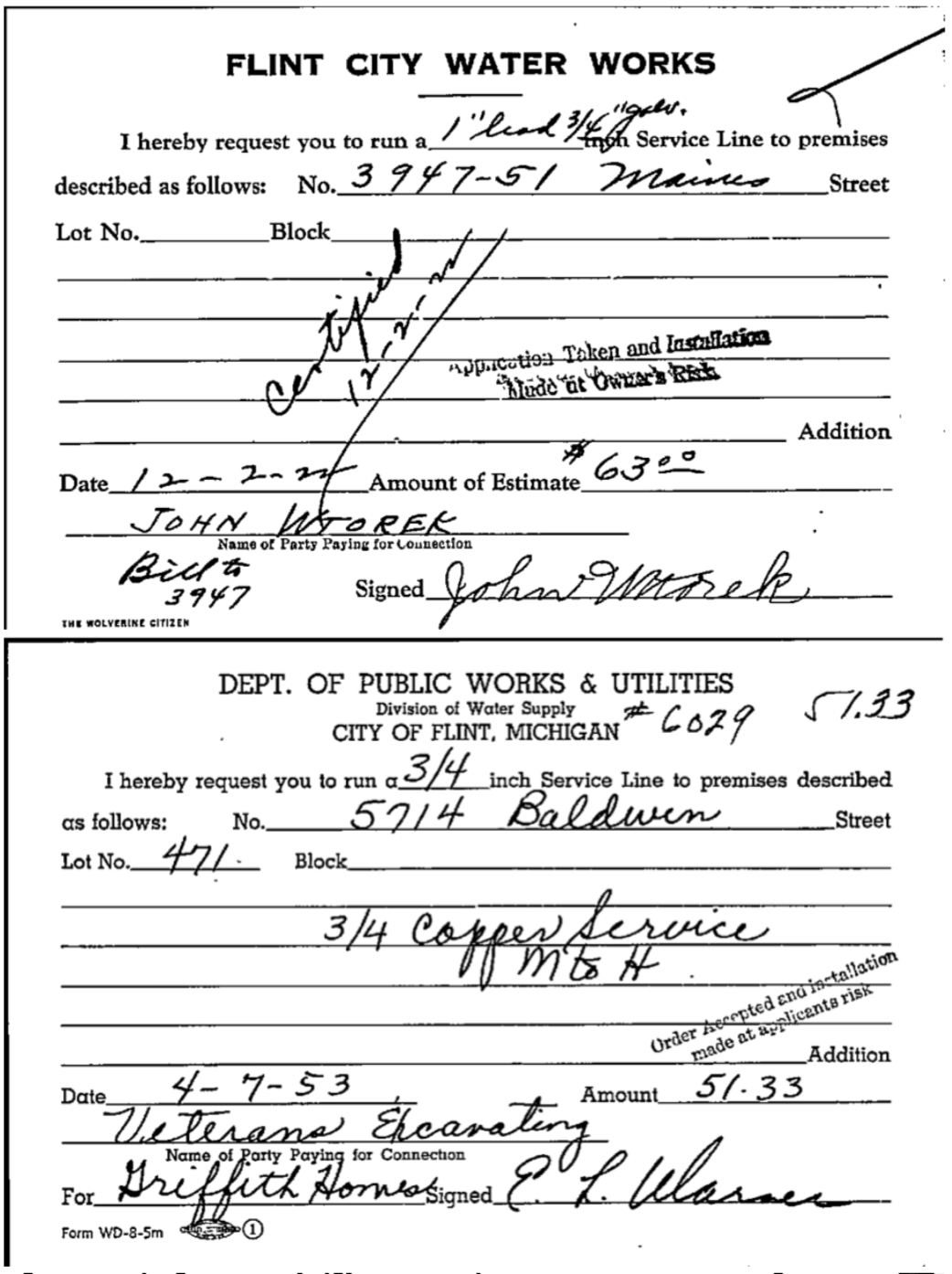}
\includegraphics[height=1.3in]{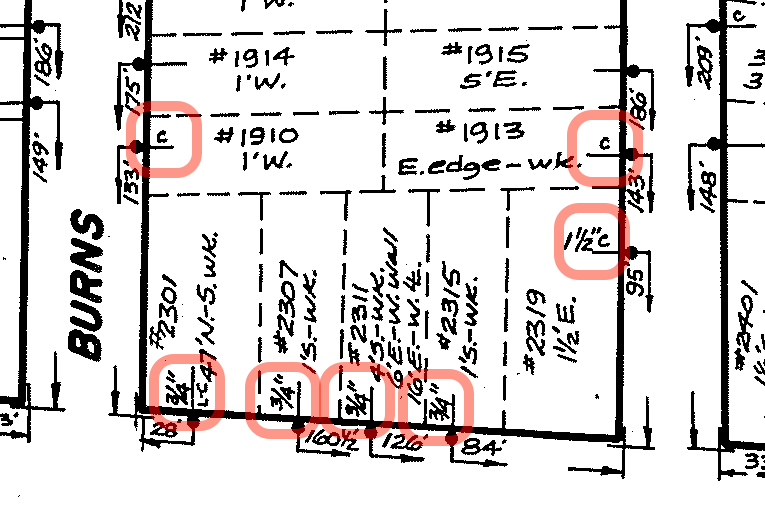}
\caption{City officials located a set of over 100,000 handwritten index cards (top) with recorded work information dating back over 100 years, and annotated maps with data on home SLs (bottom). Red circles added to emphasize markings denoting material types.} \label{fig:cityrecords}
\end{center}
\end{figure}

Initially, Flint struggled to produce any record of the materials in the city's service lines. Eventually, officials discovered a set of over 100,000 index cards in the basement of the water department\footnote{http://www.npr.org/2016/02/01/465150617/flint-begins-the-long-process-of-fixing-its-water-problem} (see top of Figure ~\ref{fig:cityrecords}). As part of a pro bono collaboration, the handwritten records have been digitized by \url{Captricity.com} and provided to the City of Flint.\footnote{We would like to thank Captricity, especially their machine learning team, Michael Zamora, Michael Zamora, David Shewfelt, and Kayla Pak for making the data accessible.} Around the same time, a set of hand-annotated maps were discovered that contained markings for each parcel that specified a record of each home's service line (bottom of Figure ~\ref{fig:cityrecords}). The map data was digitized by a group of students from the GIS Center at the University of Michigan-Flint lead by the director Prof. Martin Kaufman \cite{Mlive-Flintdataonleadwater:url}. Many of the entries in the city's records list \emph{two} materials for a given record, such as ``Copper/Lead,'' but they do not specify the precise meaning of the multiple labels. However, our latest evidence suggests that, at least in the typical case, the double records were intended to specify that the second label (``Lead'' in ``Copper/Lead'') indicates the public service line material (water main to curb stop), and the first label describes the private service line (curb stop to home), while an entry that is simply given as ``Copper'' may refer to both sections or only one. Lastly, there are a number of entries in the records that say ``Copper/?'' for the service line material, indicating missing information for the service line on the original handwritten records. Many other records are simply blank, recorded as ``Unknown/Other.''

\subsection{Peak of Crisis \& Replacement Pilot}
\label{sec:deq-in-home-inspection}

In the wake of the crisis the State of Michigan began to discuss plans for lead abatement in Flint. It had become clear to lawmakers in Michigan that they would need to invest in a large-scale removal of lead pipes from the city. To begin, FAST Start initiated a pilot phase, with the goal of replacing the service lines of a small set of residences. Flint's Mayor and the FAST Start team awarded a contract to Rowe Engineering to replace pipes at 36 homes around the city. They selected these homes based on risk factors including the presence of high water lead levels, pregnant women, and children younger than 6 years old. Nearly all of the homes, 33 of 36, had some hazardous material (lead or galvanized) in one or both portions of the service lines, while only 3 were safe. Therefore, the number of homes with physical verifications of both service line portions through September 2016 was only 36 out of over 55,000 homes. A map showing the progress of replacement in Flint can be found in Figure~\ref{fig:maps_by_time}.

\begin{table}[h]
\centering
\small{
  \begin{tabular}{|lrrrrrrr|}
       \hline
    & \multicolumn{6}{l}{Verified SL Materials (Public-Private)} &  \\
    City   Records&   C-C&   C-G&   L-C&   L-G&   L-L& Other&  All  \\ \hline
            Copper&  1115&    10&   258&    84&    13&     9&  1489 \\
         Cop./Lead&   109&    20&   816&    91&    15&    25&  1076 \\
       Galv./Other&   113&    18&   565&  1286&    81&    31&  2094 \\
              Lead&    24&     2&    29&    14&    12&     3&    84 \\
           Unknown&   152&    18&   535&  1169&   118&    42&  2034 \\
              \hline
  \end{tabular}
}
\caption{
  Discrepancies between city records of service lines, and materials verified via inspection or replacement.
}\label{tbl:sl-records-vs-truth-slr-hvi}
\end{table}

Meanwhile, in order to gather reliable information about private part of the service lines, the Michigan Department of Environmental Quality (DEQ) directed a team of officials and volunteers from the local plumbers union to personally inspect a sample of the homes of Flint residents. The public portion of the service line runs entirely under the street and sidewalk, while the private portion runs directly into the basement of the residents' home. Thus, the private portion can be inspected without any digging. The DEQ inspectors submitted their inspection results. As of June of 2016, the department had collected a data from over 3,000 home inspections.  We consider this data to be reliable, since it was curated by DEQ officials who provided it to our team. This dataset allowed us to partially evaluate the reliability of the city records discussed in Section~\ref{sec:precrisisdata}. It is important to note that the comparison is not ``apples to apples,'' as the DEQ inspections were private-portion only whereas the labels in the city records did not specify which portion of the line was indicated. We report the confusion matrix between DEQ inspection data and city records in Table ~\ref{tbl:sl-records-vs-truth-slr-hvi}. The comparison suggests that, while the records were correlated with ground truth, the discrepancies were substantial.

\subsection{Large-Scale Replacement, Mid-2016 to Now}

Our group at the University of Michigan began engaging with the FAST Start team in the summer of 2016. One of the critical decisions the team needed to make was the selection of homes that would be recommended for service line replacement. According to the FAST Start payment agreements, contractors receive roughly half (\$2500) the cost of a full replacement (\$5,000) for excavated homes with copper on both public and private portions, due to removing concrete, refilling concrete, machine use, and labor. The choice of homes was deemed critically important, as the excavation of a home's service line that discovers a ``safe'' (e.g., copper) pipe is effectively wasted money, aside from the benefit of learning of the pipe's true material. Our work has focused on minimizing such unnecessary excavations, using the tools we describe below.

\subsubsection{Early Replacement Activity and Findings (Fall 2016)}
\label{sub:replacement_phase_1_and_early_results}

By summer 2016, FAST Start had selected a set of 200 homes for replacement, scheduled to begin August, $31^{\rm st}$.  This selection is called Phase One. Like the Pilot Phase, their criteria included the presence of high water lead levels, pregnant women, children under six years old, as well as veterans and the elderly. In the present section, we describe how we helped facilitate data collection for Phase One, and how the results forced us to rethink our objectives and adjust our models. 

By late September 2016, the early data from the service line replacement program began to arrive, and the rate of lead and other hazardous pipes discovered was alarming; 96\% (165/171) of excavations revealed lead in the public portion of the line. These findings differed significantly from the city records, which had previously indicated that among those homes only 40\% would contain lead in either portion. As data from Phase One arrived it was becomingly increasingly clear that \emph{likely over 20,000 homes} have unsafe pipes serving their water -- dramatically higher than earlier estimates. Critically, as these discoveries were being made, a debate was taking place in the U.S. Congress discussing the possibility of more than \$100M in funding for the Flint's recovery efforts.

With the debate in the Congress ongoing, our team decided to put out an informal report to raise the alarm about the extent of the lead issue, and several news outlets reported on our findings \citep[e.g.][]{MRadio-Flintmighthave:url,FarmoreFlinthomes:url}. This effort lead to a formal report in November of 2016 that provided a more precise estimate of the number of lead replacements likely to be needed \citep{CityOfFlint:url}, which was provided to the city's mayor, the DEQ, and the U.S. Environmental Protection Agency. Our report, based on comparing the city records and the data gathered from contractors, suggested that the number of needed replacements would be between 20,600 and 37,100. The large range accounts for the inherent uncertainty in data collection and model assumptions, as well as the question of \emph{occupancy}. One challenge that is specific to Flint is the fact that around one third of the city's homes are not occupied, a rate that is the \emph{highest in the country}\footnote{\url{https://www.reuters.com/article/us-flint-vacancies-idUSKCN0VK08L}}.

\subsubsection{Contractor Data Collection Application} \label{sec:collection}

With thousands of homes scheduled to have their water service lines excavated by multiple contractors, the collection and management of the data generated by this large-scale effort would prove to be a logistical challenge.  While initially there was a plan in place to collect data via paper forms that would later get transferred to a spreadsheet, it was increasingly clear that digitally recording information, and storing it centrally, would be a more effective strategy and less prone to error.

 \begin{figure}[!ht]
 \begin{center}
 \includegraphics[width=2.5in]{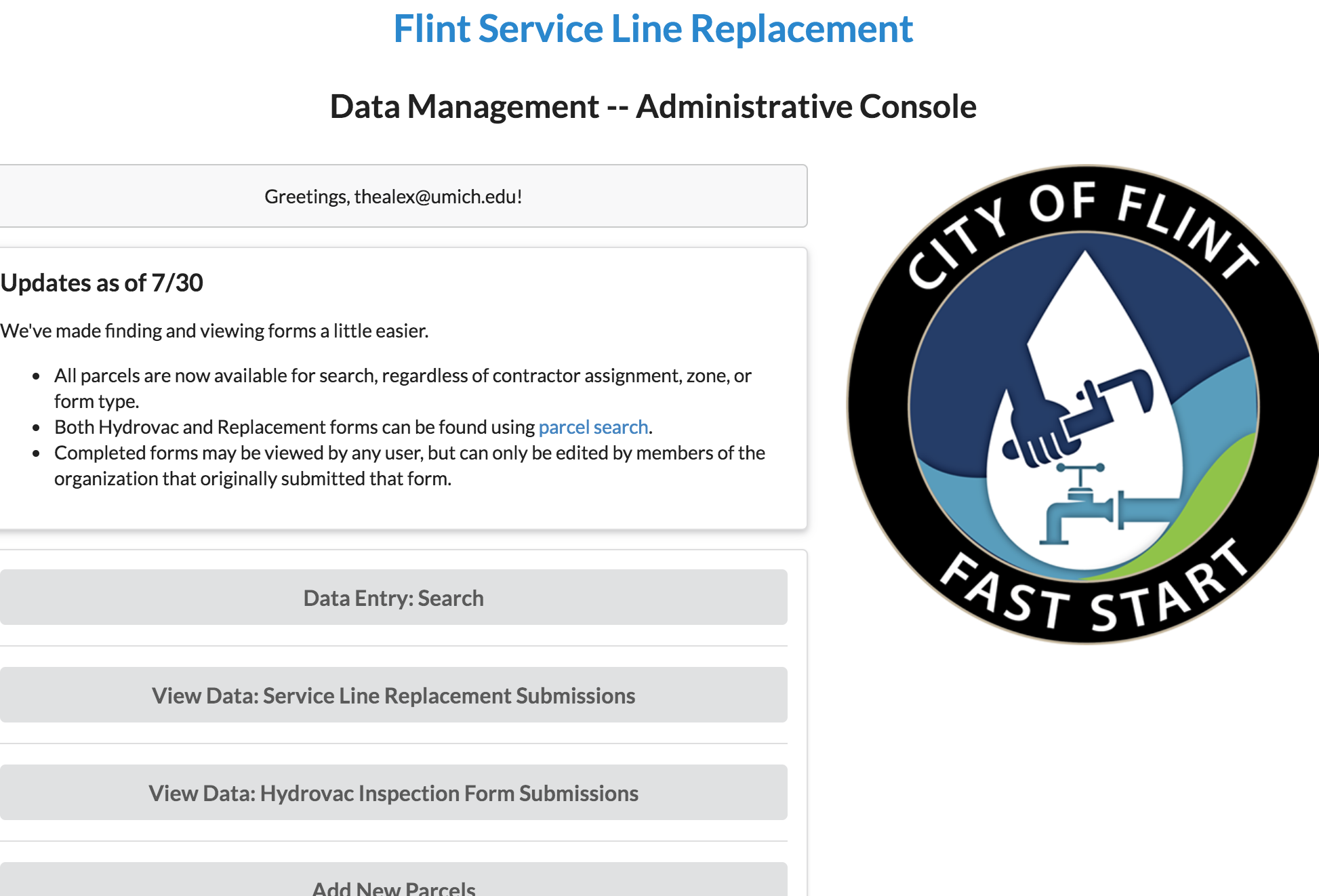}
     \caption{Mobile and web app, developed by the authors, to gather replacement data from contractors on-site.}
     \label{fig:app_sl_collection}
 \end{center}
\end{figure}

Our team volunteered to facilitate the data collection efforts. In the fall of 2016, we developed a web and mobile application with various access levels. The latest version of this app is a custom-built web application using Python and the web framework using Flask. The users, on-site contractors as well as DEQ and Fast Start officials, are asked to select homes and to fill in essential information about service line work accomplished at each site. This information includes the excavated pipe materials, lengths, dates, and data on the home's residents. The output of the form appears in real-time in a live database with mapping capabilities. We adopted a tiered permissions structure with password-protected information to maintain the privacy of the data. The app continues to be used as of this writing for tracking progress for the public and for paying contractors for completed work.

\subsubsection{Hydrovac Digging: Inspection without Replacement}
\label{sec:unbiased_data_and_the_hydrovac_pilot}

The foremost challenge of a large-scale service line replacement program is the uncertainty about which homes possess safe service lines and which homes have lines made of hazardous materials. As of the summer of 2016, the only concrete verified data on pipe materials across the city consisted of the 36 data points provided by the Rowe engineering. By the end of Phase One, this number increased to about 250 homes. At this point, the excavation of pipes at a single home would cost anywhere from \$2,500-5,000, a prohibitively high cost for data collection. At the same time, the available replacement data consisted of \emph{cherry-picked homes}: houses were selected for line replacement if they were presumed to have an overwhelming likelihood of lead. These addresses and were highly concentrated in only three neighborhoods (see Figure~\ref{fig:maps_by_time}) and provided nothing close to a representative sample of the broader city. We therefore realized, and emphasized to members of FAST Start, that the effort required a cheaper, quicker, and more statistically sound method to gather data.

\begin{figure}[!ht]
\begin{center}
\includegraphics[height=1.1in]{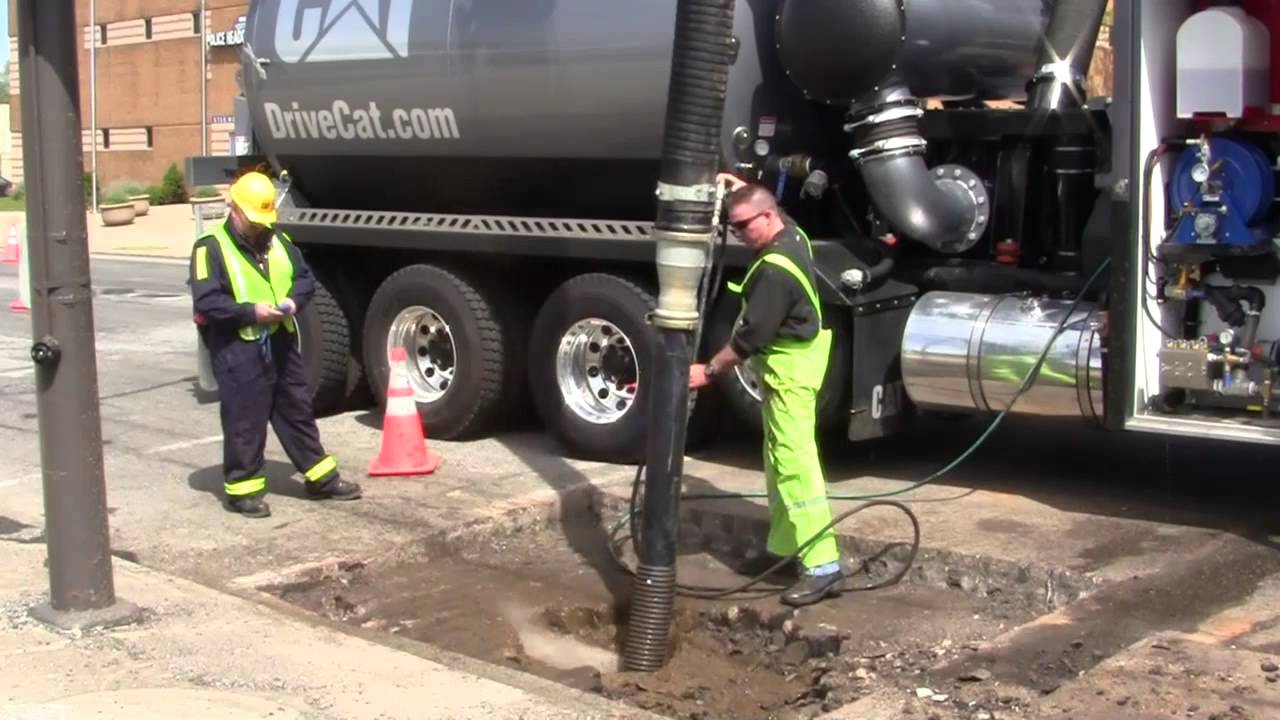}
\includegraphics[height=1.1in]{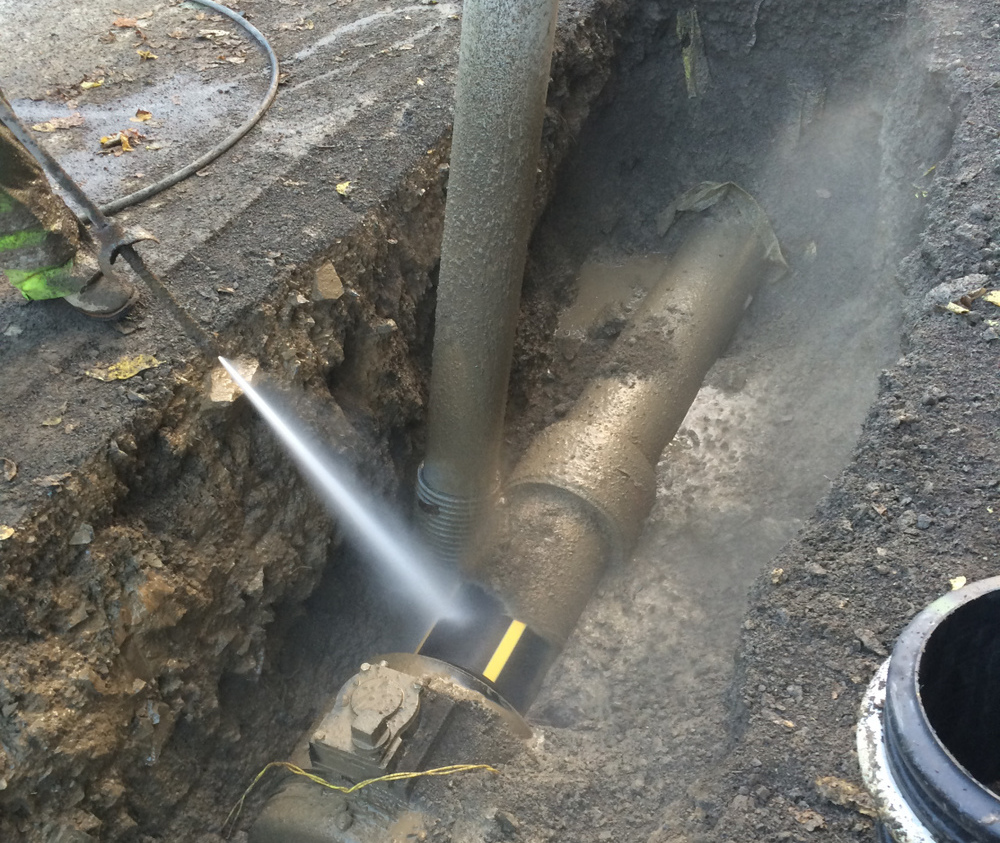}
	\caption{Using a hydrovac truck for inspection, requires a large truck and crew (left) and exposes the pipe material underground (right).
  }
	\label{fig:hydrovac}
\end{center}
\end{figure}

After a lengthy discussion with water infrastructure experts and contractors, a new alternative emerged: \emph{hydrovac inspections}. A hydro-vacuum truck, or simply a hydrovac (see Figure~\ref{fig:hydrovac}), has two main components: a high-pressure jet of water used to loosen soil and a powerful vacuum hose that sucks the loosened material into a holding tank. The hydrovac technique allows workers to dig a small hole quickly and then inspect whatever is observed underground. It is ideal for determining service line materials, as it can dig at the location of the home's curb box (connects the home's service line at the property line to the water main), and observe the pipe materials for both the public and private portions of the service line. The cost can be as low as \$250 per inspection and often does not require prior approval from residents, as the digging site is mostly confined to city property. One limitation is that the hydrovac can only dig through the soil, and not through driveway or sidewalk pavement. This limitation led to unsuccessful excavations 20\%-25\% of the time, according to the hydrovac engineers.

The selection of homes for hydrovac inspection was one of the primary contributions of our team to FAST Start's efforts, and we were given wide discretion for ``sampling'' homes. This reflects the political and logistical challenges of service line replacement, as full excavation of service lines required a much longer process with oversight by the city council. We would emphasize that, in the following section where we describe our sequential decision protocols, our primary focus was on the model and inspection subroutines, and we assume the replacements are made using a simple greedy strategy.

\section{Prediction \& Decision Framework}
\label{sec:overall_framework}

In this section, we formally define the sequential decision-making problem for a city, in our case the city of Flint, seeking to remove all of the lead service lines from its homes under the following conditions:
(i) for almost all homes, the service line materials of homes are unknown;
(ii) there is a method of inspection to collect information;
(iii) it is costly to excavate service lines that do not need to be replaced; and
(iv) there is a fixed budget for replacement and inspection.

There are $N$ total homes in the city, and it is unknown which homes need new service lines. We let the unknown label for home $i$ be $y_i \in \{0,1\}$, taking on the value 1 if the home needs a replacement and 0 otherwise.  Note that a home needs replacement if either the public \emph{or} private portion of the service line is hazardous. We also have information about each home, denoted by a vector $x_i$, with $m$ features, that describe it (see Section \ref{sec:data_availability}).  We want to learn the label $y_i$ given $x_i$, for each $i=1,\ldots,N$.  We divide the procedure to find out these labels into two steps: first, a statistical model for prediction (\statmodel); and second, an algorithm that decides which homes to observe next (\druleHVI).

There is another decision rule, \druleSLR, that determines which pipes to replace next. \druleSLR\ is a \emph{greedy} algorithm. That is this algorithm recommends that the replacement crew should go to the homes with the highest probabilities of having hazardous pipes. Given that, our \druleHVI\ is focused on learning, and \druleSLR\ uses that learning to reduce costs. 

\subsection{\statmodel}
\label{sec:model_for_prediction}

In this section, we describe \statmodel, which assign a probability that a service line contains hazardous materials.  \statmodel\  is a novel combination of predictive modeling using machine learning and Bayesian data analysis.  First, a machine learning prediction model gives a prediction for the public and private portion of each home's service line using known features.  These predictions then become the parameters to prior distributions in a hierarchical Bayesian model designed to correct some of the limitations to the machine learning model.

\subsubsection{Machine Learning Layer}

The machine learning layer of \statmodel\ outputs a probability of having a hazardous service line material for each home for which the material is unknown. Specifically, this layer gives a prediction, $\hat y_{i,k} = f_\theta(X_{i,k})$,
the probability that service line portion $k$ for home $i$ is hazardous, and $X_{i,k}$ is a vector of features, described in Section \ref{sec:precrisisdata}. After examining several models empirically (see Section \ref{sec:empirical_prediction}) we chose the machine learning layer, $f_{\theta}()$, to be XGBoost, a boosted ensemble of classification trees \cite{chen2016xgboost}.

\subsubsection{Hierarchical Bayesian Spatial Model Layer} \label{sec:HBayesModel}

One limitation of classification algorithms is how they handle unobserved variables, which may be correlated with the outcome. We address this limitation with a hierarchical Bayesian spatial model. This accounts for unobserved heterogeneity related to geographic location and similiarity of homes, which is used in hierarchical spatial models with conditional autoregressive structure \citep{gelman2014bayesian,gelfand2003proper,lee2011comparison,lee2013carbayes}. Empirically, each geographic region across the city (e.g., voting precincts) has a different number of observed service lines. While a city-level (pooled) model ignores precinct differences and a separate (unpooled) model for each precinct is limited by small sample sizes or even no observations, our full hierarchical (partially pooled) model strikes a balance with shrinkage. Precincts with little information will have their parameters pulled towards the city-wide distribution. Details of the Bayesian model, and how these are combined with the machine learning layer, are explained further in the full version of the paper.

\subsection{\druleHVI}
\label{sec:alg_for_selecting}

Now we describe \druleHVI, which utilizes active learning \cite{balcan2013statistical,balcan2010true,liu2008active} to efficiently allocate scarce resources to find and replace hazardous service lines. In general, a decision-maker may choose any active learning algorithm for inspection. In this work, we implement a version of Importance Weighted Active Learning (IWAL).

\begin{table}
	\caption{Summary of notation }\label{tbl:notation}

	\begin{tabular}{cl}
	\hline
	Notation & Explanation \\
	\hline
	$\mathcal{X}$ & observable feature space for each parcel/home \\
  $x_i, y_i$ & observable features for home $i$, label for home $i$ \\
	$\inspect_t/\replace_t$ & indicates ``home $i$ inspected/replaced at $t$?''\\
	$y_t^\inspect/ y_t^\replace$ & indicates ``learned $i$'s' label via inspect./replace?''\\
	$Q_{it}$ & indicates ``learned $i$'s label at $t$?''\\
	$q_{it}$ & indicates ``already know $i$'s label at $t$?''\\
	$c^h, C^{\replace+}, C^{\replace-}$ & cost of inspect., successful SLR, \& failed SLR \\
	$U_t, L_t$ & set of labeled/unlabeled data at $t$\\
	\hline
	\end{tabular}
\end{table}

\subsubsection{Active Learning Setup: Inspection and Replacement}

We begin by describing the problem of efficiently locating and replacing hazardous pipes in a pool-based active learning framework (see Algorithm \ref{alg:al_mab}). Consider a budget of $B$ total queries and a pool $\pool = \left \{ x_1,\ldots,x_n \right \}$ of unlabeled homes.  Then at each time period $t$ the algorithm will produce a probability vector $\phi_t = (\phi_{1,t},\ldots,\phi_{n,t})$ that gives the probability that any home $i$ is chosen at $t$.

Contractors can determine the material of a service line via either hydrovac inspection or service line replacement. When home $i$ is chosen for hydrovac inspection at time $t$, we denote $\inspect_t = i$.  When the service line for home $i$ is replaced at time $t$, we denote $\replace_t = i$.  Once inspected or replaced, $y_i$ is known for all subsequent rounds $t, t+1, \ldots $ and $p_{i,k}$ becomes 1 or 0, and we define $q_{i,t}=1$ if home $i$ has been observed through round $t$. $n_t^{\inspect}$ and $n_t^{\replace}$ are the number of hydrovac and replacement visits, respectively. The number of successful replacements is denoted as $n_t^{\replace+}$ (true positives) and the number of unnecessary replacements as $n_t^{\replace-}$ (false positives).

We initially set $U_0 = \pool$, and let $U_t = \left\{ x_i | q_{i,t} = 0 \right\}$ be the set of homes whose service line material is unknown at time $t$, and $L_t$ be the set of homes with known service line materials.  Finally, the budget also allows for a fixed number of inspections $d$ for each period. The problem is how to select these $d$ homes with unknown labels at each period $t$ to maximize information gained.

\begin{algorithm}
\caption{\activerem\ for MultiEpochReplacement sequentially selects homes for both inspection and for replacement each epoch, incorporating ideas from both active learning and multi-armed bandits.}\label{alg:al_mab}
\footnotesize{
\begin{algorithmic}[1]
\State \textbf{Input}: parameters $B,N,T$
\State \textbf{Input}: initial observed data
\For{$t = 1,...,T:$}
  \State Update: $\hat{p}_t(\theta)
    \gets \statmodel\ p_t(\theta, L_{t-1})$
  \State Inspect:
    $\inspect_t \gets\druleHVI\ \phi^{\inspect} (\theta)$
  \State Observe labels: $y_t^{\inspect}$
  \State Update: $\hat{p}_t(\theta)
    \gets \statmodel\ p_t(\theta, \{ L_{t-1}, y_t^{\inspect} \} )$
  \State Replace:
    $\replace_t \gets\druleSLR\ \phi^{\replace} (\theta)$
  \State Observe labels: $y_t^{\replace}$
  \State Update: $U_t \gets \{U_t \} \setminus \{ \inspect_t,\replace_t \}$,  $L_t \gets \{L_t \} \cup \{\inspect_t, \replace_t \}$
  \State $\text{TotalCosts}_t \gets \text{TotalCosts}_{t-1} + ( c^h + \mathbf{1}^{\replace+}_{(c^{\replace+})} + \mathbf{1}^{\replace-}_{( c^{\replace-} )} ) $
\If { $\text{TotalCosts}_t \leq B$ check budget} continue \Else { stop}
\EndIf
\EndFor
 \State $\text{HitRate}_T^{\replace} \gets n_{T}^{\replace+} / (n_{T}^{\replace+} + n_{T}^{\replace-})$
\State $\text{EffectiveCost}_T = \text{TotalCosts}_T / n_{T}^{\replace+}$
\end{algorithmic}
}
\end{algorithm}

\subsubsection{Simple Active Learning Heuristics: Uniform and Greedy}

We first propose several benchmark strategies for selecting homes for inspection. This family of algorithms randomly alternate between \emph{random exploration} of the unobserved data and \emph{greedy inspection} of the highest-predicted hazardous homes. As we see in Table \ref{tbl:decision_rules}, these decision rules differ in the costs they incur.
\begin{itemize}
\item \textbf{HVI uniform} \emph{(egreedy(1.0))}: Select homes uniformly at random from the pool of those with unknown service lines.
\item \textbf{HVI greedy} \emph{(egreedy(0.0))} Select the homes most likely to have hazardous service lines, based on current model estimates.
\item \textbf{HVI $\varepsilon$-greedy}  \emph{(egreedy($\varepsilon$))}:  For a $1-\varepsilon$ fraction of the inspections, select \emph{greedily}, that is select homes for HVI based on the highest predicted likelihood of danger. For the remaining $\varepsilon$ fraction, select homes uniformly at random for HVI.  We experiment with values $\varepsilon = \{0.1,0.3,0.5\}$. Also, we note that \textbf{HVI uniform} and \textbf{HVI greedy} are special cases, with $\varepsilon$ set to $1.0$ and $0.0$, respectively.
\end{itemize}

\subsubsection{Importance Weighted Active Learning}

We propose an algorithm that takes in the current beliefs about whether each home has hazardous pipe material, and outputs a decision of which homes should be inspected next period. This proposal is a variant of the Importance Weighted Active Learning (IWAL) algorithm \citep{beygelzimer2009importance}. The key idea behind IWAL is to sample unlabelled data from a \emph{biased} distribution, with more weighted placed on examples with greater uncertainty, and then after obtaining the desired labels to incorporate the new date on the next iteration of model training. Our implementation of this approach takes the part of \druleHVI\ which is core to Algorithm \ref{alg:al_mab}. A full explanation of our \textsc{IWAL} implementation will be available in the full version of the paper.

\subsubsection{Analyzing Costs}

There are two categories of costs incurred in Algorithm \ref{alg:al_mab}: hydrovac inspections and replacement visits. Hydrovac inspections always cost the same amount and are denoted $c^\inspect$. Service line replacement costs, however, depend on what is actually in the ground. If contractors excavate a service line that does not need be replaced, we still incur a cost $c^{\replace-}$ for labor and equipment, even though no replacement occurred. On the other hand, if contractors uncover a line that needs to be replaced then the direct cost of replacement is $c^{\replace+}$.

But \emph{effective cost per successful replacement} is greater than its direct cost, and we define formally it as $\text{TotalCosts} / n^{\replace+}$, where
\[
\text{TotalCosts} = c^{\inspect} n^{\inspect} + c^{\replace+} n^{\replace+} + c^{\replace-} n^{\replace-}
\]
(See Algorithm \ref{alg:al_mab}). In Flint, hydrovac inspection costs are summarized in Table \ref{tbl:decision_rules}. We note that the effective cost of a successful replacement is driven by two factors: the model accuracy ($\text{HitRate}^{\replace}$) and the ratio of their costs, $c^{\replace-}/c^{\inspect}$. Since unnecessary replacement visits can be avoided by prior inspection with a hydrovac, these two metrics, which naturally vary by city, will be critical guides to applying this approach to other cities. 

\begin{table}[!t]
\centering
\scriptsize{
  \begin{tabular}{llrrrr}
  \hline
   & & 					&  	\multicolumn{3}{c}{Homes visited by rule} \\
  Hydrovac Inspection & Replacement Visit & (Cost) & \emph{Uniform} & \emph{None} & \emph{10\%} \\
   \hline
  Finds Safety & $\to$ not needed  & (\$250) 		& 230 	 & 0 	 & 23 \\
  Finds Danger & $\to$ replaces Danger & (\$5,250) 	& 770 	 & 0 	 & 77 \\
  None & $\to$ finds Safety & (\$2,500) 				& 0 	 & 230 	 & 207 \\
  None & $\to$ replaces Danger & (\$5,000) 				& 0 	 & 770   & 693 \\
  \multicolumn{3}{r}{Effective Cost per Successful Replacement:} 		&\$5,325 &\$5,747&\$5,705 \\
  \hline
  \end{tabular}
  \caption{Average effective cost per successful replacement varies by simple \druleHVI, shown by 1,000 home visits.}\label{tbl:decision_rules}
}
\end{table}

\section{An Empirical Analysis in Flint}

\label{sec:empirical_flint}

In our empirical analysis, we use the data of the confirmed service line material from the 6,505 homes identified and replaced by Flint FAST Start, as of September 30, 2017 collected via our data collection app. This data is combined with our supplementary datasets describing homes (Section \ref{sec:data_availability}) and we train a suite of classification models to predict the presence of hazardous service line materials for a given home, and the predictive power of each model is measured on hold-out sets of homes (Section \ref{sec:empirical_prediction}). After selecting a strong empirical model, we utilize the model predictions in our decision-making algorithms, which recommend those homes which will be most informative for inspection, and also those most likely containing hazardous service line materials for replacement (Section \ref{sec:empirical_optimization}).

We emphasize that our methods and models were utilized by FAST Start officials for the management of the hydrovac process, and during the early days of the efforts we were given discretion over which homes would receive inspections. We used this freedom to select statistically representative samples, as well as targeted inspections on homes of interest. In practice, our modeling efforts had less impact on the choice of replacement homes, as these decisions carried greater political and logistical challenges.

\subsection{Classification Algorithm Performance}
\label{sec:empirical_prediction}

Selecting a robust, precise, unbiased, and properly calibrated classification algorithm is key for our proposed active learning framework. Ultimately, the selected decision-making algorithm requires both accurate and well-calibrated probability estimates when selecting the next round of homes to investigate. To select such a classification model, we employ several machine learning model and compare them across various performance metrics. These metrics include the Area Under Receiver Operating Characteristic curve (AUROC), learning curves, and confusion matrices (including accuracy and precision). Using these scores, we find that tree-based methods are the most successful and robust category of models for this data. In particular, the model for gradient boosted trees implemented in the package \texttt{XGBoost} exhibits the strongest performance with a fewest data points.

\subsubsection{ROC and Learning Curves}

\begin{figure}[!ht]
\centering
\includegraphics[width=0.7\linewidth]{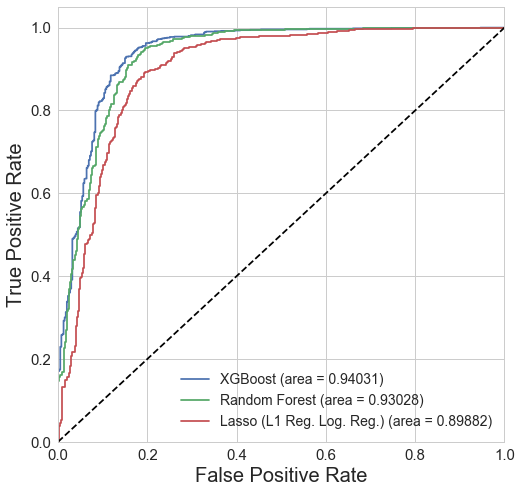}
\caption{ROC curves measuring predictions of XGBoost, RandomForest, and lasso logistic regression on a random holdout set of all available data.}
\label{fig:roc_curve_one_run}
\end{figure}

\begin{figure}[!ht]
\centering
\includegraphics[width=0.7\linewidth]{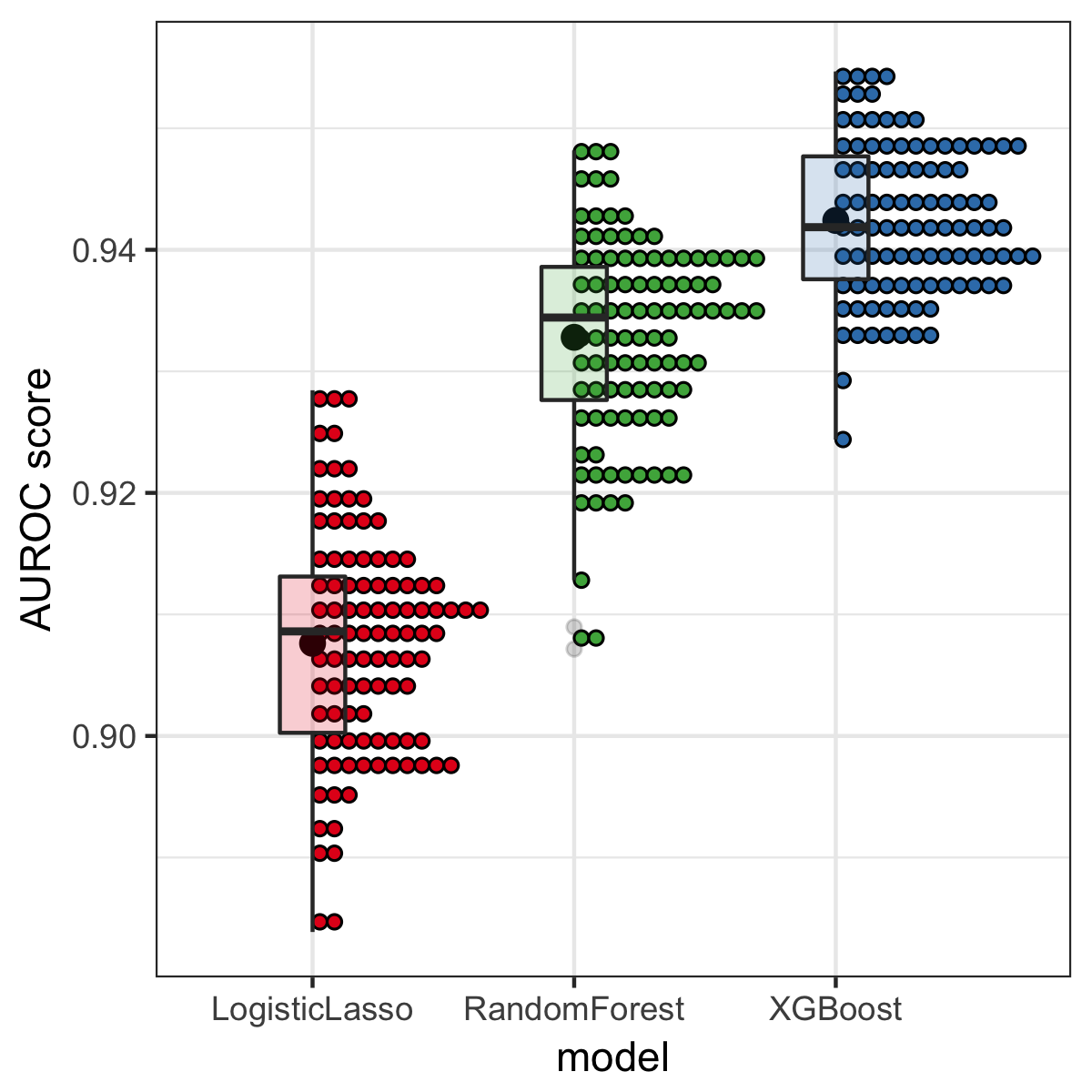}
	\caption{Empirical distributions of AUROC scores of classifiers over several runs on random holdout sets.  Both XGBoost and RandomForest show marked performance improvement over lasso logistic regression, and XGBoost gives marginal improvement on RandomForest.} \label{fig:roc_scores_classifiers}
\end{figure}

\begin{figure}[!t]
\centering
 \includegraphics[width=0.8\linewidth]{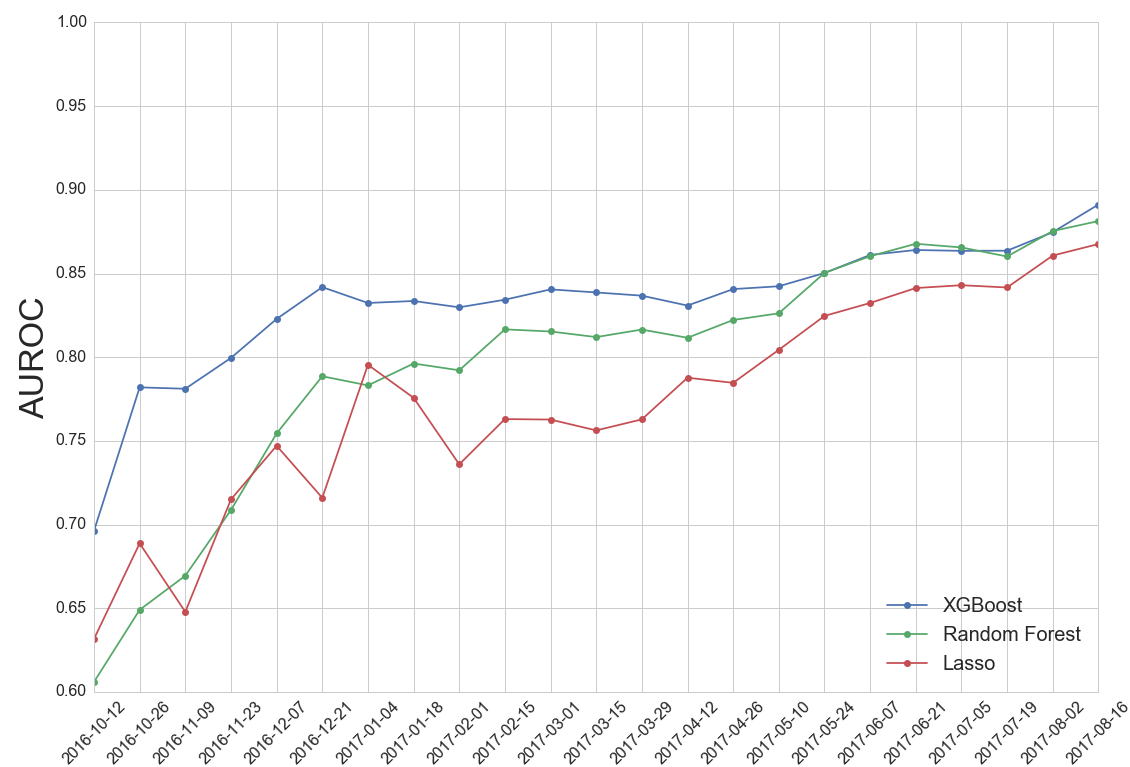} 
	\caption{Temporal Learning curves for classification of hazardous service line materials. XGBoost consistently outperforms the other classifiers, especially at the beginning of the timeline when there is less data available.}
	\label{fig:learn_curve}
\end{figure}

The overall accuracy of the best performing XGBoost model, based on a holdout set of 1,606 homes (25\% of available data), is 91.6\%, with a false-positive rate of 3\% and false-negative rate of 27\%. The homes falling in the top 81\% of predicted probabilities are classified as having hazardous service lines. The ROC curves and AUROC scores show XGBoost's superior performance with an AUROC score of 0.939 on average in a range of [0.925, 0.951], Figure \ref{fig:roc_curve_one_run} and \ref{fig:roc_scores_classifiers}). While the ROC curves show a single run of each model, the AUROC scores are shown as distributions of 100 bootstrapped samples obtained using a stratified cross-validation strategy with 75\%/25\% of the data randomly selected for training/validation. We further examine AUROC scores using learning curves (Figure \ref{fig:learn_curve}), using random subsets of data to illustrate diminishing returns of additional data on model performance using AUROC.  We also introduce, \emph{temporal learning curves}. These temporal learning curves reflect the exact order of data collection in 2016-17, and they show the AUROC as we re-estimate the model every two-week period to predict the danger for all remaining not-yet-visited homes. We finally ensure that the model's predicted probabilities, which we use to quantify our prediction uncertainty, are indeed well-calibrated probabilities. \footnote{While not shown here, we also considered ExtraTrees, AdaBoost (with decision tree classifier), and Ridge Regression (regularized with L2 loss), but performance was lower than the three presented. Full details on hyperparameter optimization will be available in the full version.}

\subsubsection{Risk factors}

Now that we have a robust predictive model, we can look at which features of a home and its surrounding neighborhood are the most predictive feature in identifying homes with hazardous service lines. But we are cautious to not make any causal claims from this analysis. We obtain the feature importance values\footnote{We calculate feature importance by weight, which is the normalized frequency with which a feature appears in a tree amongst the ensemble.} produced by each model by training with 20 bootstrapped samples of the data and reported the average feature importance values. The most informative home features relate to its \emph{age}, \emph{value}, and \emph{location}, suggesting that the context (place and time) in which the home was built, as expected, is strongly correlated with service line material. For instance, homes built during and before World War II and those that are lower in value are more likely to contain lead in their public service line. Two additional features were the \emph{city records} and the \emph{DEQ private SL inspection reports}. Each was shown to be a noisy but useful predictor, as indicated earlier in Table \ref{tbl:sl-records-vs-truth-slr-hvi}.

\subsection{\activerem: Evaluation}
\label{sec:empirical_optimization}

\newcommand{\numInitObservedHomes}{6,506\xspace}
\newcommand{\ActualFlint}{\textsc{ActualFlint}\xspace}
\newcommand{\SimulatedFlint}{\textsc{SimulatedFlint}\xspace}

We now discuss our implementation of the \activerem\ framework applied to the particular case of Flint's large-scale pipe replacement program. With over \$100M in investment, Flint is a perfect testbed to compare the performance of our proposed methods (developed in Section \ref{sec:alg_for_selecting}) with the actual empirical performance of the work of FAST Start thus far. Our goal is to show a high potential for savings by minimizing the number of unnecessary replacement visits, thus replacing more hazardous lines under the same budget.

\subsubsection{Experimental testbed, and potential biases.} Any experimental framework needs a quality dataset, with known labels for a large sample which we can evaluate our procedure. Fortunately for the City of Flint, where contractors have been working for over 18 months, we have a total of \numInitObservedHomes\ observations of service line materials. A natural choice for an experimental environment, which we call \ActualFlint, is to use the set of observed homes in Flint as a template for the overall city, i.e. a municipality with precisely \numInitObservedHomes\ homes whose service line material we can query as needed.

A major challenge of relying solely on observed data is that the actual home selection process is biased, in both the hydrovac inspections and the line replacements. While a certain fraction of the home selection was random,  it was often reasonably arbitrary due to political and logistical constraints. For instance, many of the homes selected for service line replacement were chosen to maximize lead discovery. To assess the effect of sample bias, we developed an experimental environment, \SimulatedFlint, in which we suppose Flint contains only those properties \emph{not in} the observed dataset. For this dataset, labels are assigned based on the labeled hold-out data. With observed data as training, we used a K-Nearest-Neighbors (KNN) classifier to estimate a probability for each unknown home, and then sampled a Bernoulli random variable -- "safe"/"unsafe" -- to assign labels. This randomized dataset has lower potential selection bias concerns. In the reported results below, we focus on \ActualFlint, but we note that results from \SimulatedFlint were nearly equivalent.

\subsubsection{Backtesting Simulation on \ActualFlint}

We quantify the cost savings from implementing our algorithm by comparing the sequential selection of homes from the proposed decision rules to what the Flint FAST Start initiative actually did in 2016-17. 
The goal is to stretch the allocated funds to remove hazardous pipes from as many homes as possible. One source of inefficiency in spending is unnecessary service line replacement (SLR) visits (the false-positive error rate). Therefore, our key performance metric is the SLR hit rate, i.e. the percentage of homes visited for replacement that required replacement.

\begin{figure}[!t]
\begin{center}
\includegraphics[width=.7\linewidth]{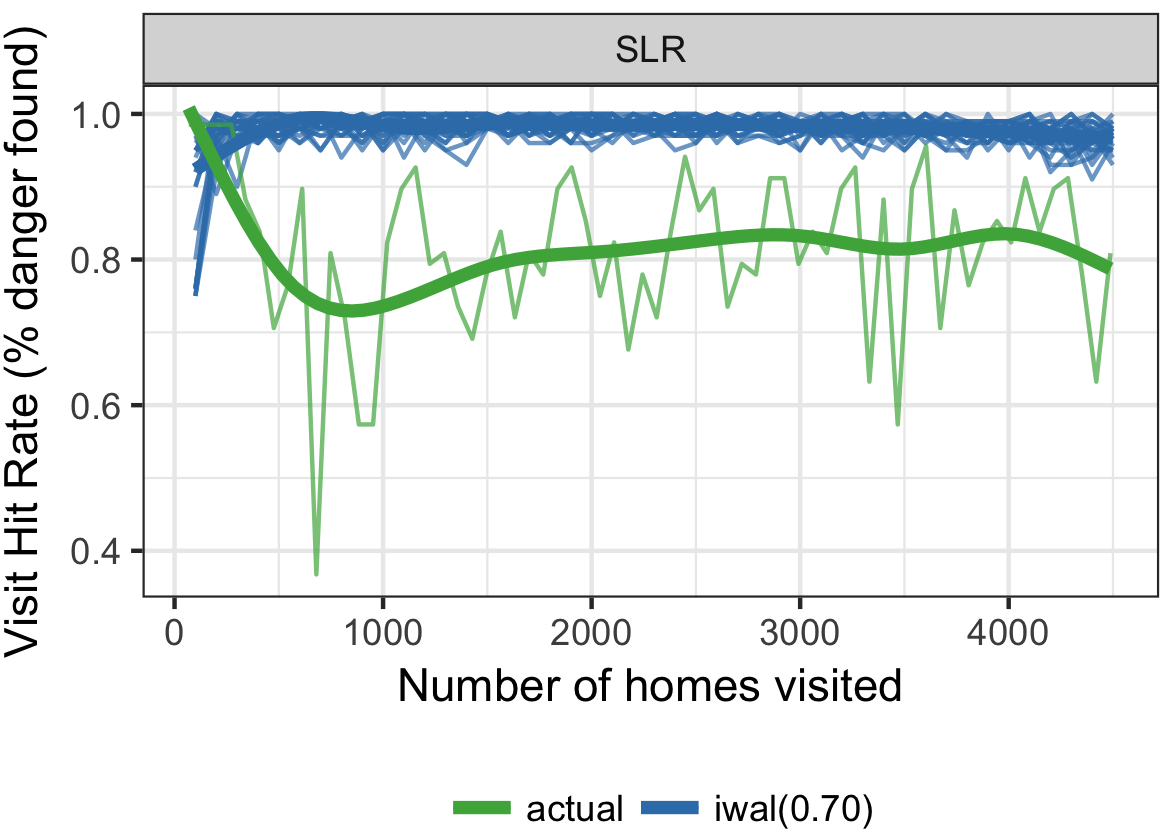}
\end{center}
\caption{
	Tracking hit rates over time, the proposed IWAL algorithm (blue; mean = 98.0\%) outperform actual (green; mean = 81.2\%; thick line is smoothed plot)
}
\label{fig:backtesting-100-even-ffs-vs-iwal}
\end{figure}

\emph{The proposed approach greatly improves the hit rate.} Our key finding from the simulation shows that we predict a reduced rate of costly unnecessary replacements visits from 18.8\% (actual) to 2.0\% (proposed). Figure \ref{fig:backtesting-100-even-ffs-vs-iwal} illustrates the direct comparison of hit rates for our proposed approach, \textsc{IWAL}(0.7), based on our \ActualFlint simulation, compared to Flint FAST Start. 

\emph{Second, the cost savings are substantial.} The proposed algorithm, with a higher hit rate, increases the number of homes that receive service line replacements for the same number of visits. This, in turn, reduces the \emph{effective cost} of a successful service line replacement. The effective cost includes both the direct costs of successful replacement visit and the average costs incurred by exploring homes from hydrovac inspections or unnecessary replacement visits. Having access to the exact same set of 6,505 homes actually observed, we find that the algorithm on average saves an additional 10.7\% in funds per successful replacement (see Table \ref{tbl:costs}). Across 18,000 total planned service line replacements, this would extend to an expected savings of about \$11M out of current spending. In terms of the overall removal of lead pipes, this is approximately equivalent to 2,100 additional homes in the city that would receive safe water lines. These estimates are made using the current costs in Flint, where hydrovac inspection costs $c^h=\$250$, unnecessary replacement costs $c^{\replace-}=\$2,500$, and successful replacement costs $c^{\replace+}=\$5,000$.

\begin{table}
\centering
\scriptsize{
  \begin{tabular}{llccc}
  \hline
  						&	& \textbf{Actual}   	& \multicolumn{2}{c}{\textbf{Proposed Algorithm\quad\quad. }} \\
  						&	&  			& Mean	& \emph{Range} \\
  \hline
  \multicolumn{3}{l}{For every 1 successful replacement:} & &  \\
  & Effective cost 						& \$5,818 	& \$5,196		& (\$5,186 to \$5,208) \\
  & Predicted savings (\$)   				& -- 		& \$621.7		& (\$610.4 to \$632.4) \\
  & Predicted savings (\%)  				& -- 	 	& 10.7\% 		& (10.5\% to 10.9\%) \\
  \multicolumn{3}{l}{For every 1,000 successful replacements, the savings generate:} & &  \\
  & Extra inspections 			& -- 		& 94			& (92 to 96) \\
  & Extra replacements 			& --		& 120 			& (117 to 122) \\
  \multicolumn{3}{l}{For 18,000 successful replacement:} & & \\
  & Predicted savings (\$ in millions) & -- 		& \$11.18m  	& (\$10.99m to \$11.39m) \\
  \hline
  \end{tabular}
  \caption{Cost savings. The proposed method lowers the effective cost per successful service line replacement, saving \$621.7 per home (10.7\%), enough to remove lead from an additional 2,000 homes on the same budget.
  } \label{tbl:costs}
}
\end{table}

\begin{figure}[!ht]
\begin{center}
\includegraphics[width=.7\linewidth]{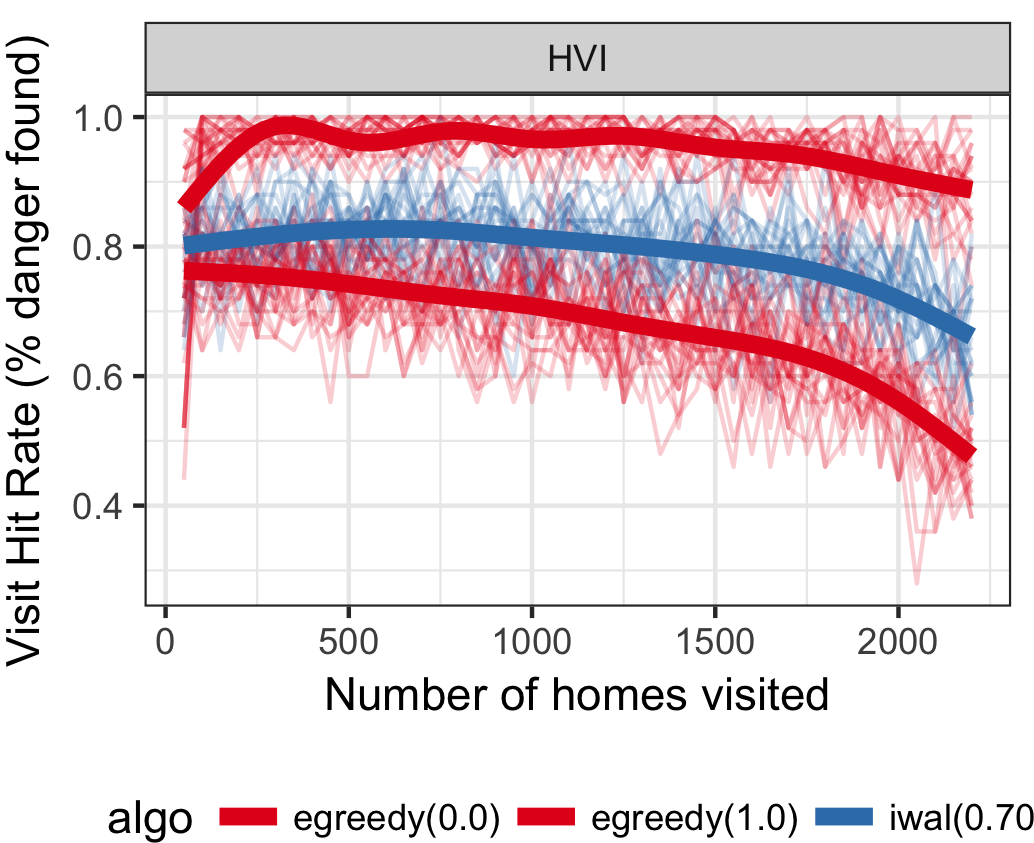}
\end{center}
\caption{HVI Hit Rates. egreedy(0) tends to over-inspect whereas egreedy(1) is too conservative. IWAL more effectively optimizes HVI hit rate.}
\label{fig:backtesting-100-even-hr-bytime-greedy-iwal-unif}
\end{figure}

\emph{The proposed approach outperforms a competitive set of natural benchmark strategies.} Instead of only comparing our proposed method to what actually occurred, we also consider a range of alternative methods. In particular, greedy (egreedy with 0\% exploration) inspects the highest rate of hazardous homes inspected (HVI hitrate 91\%), and uniform (egreedy with 100\% exploration) inspects the lowest (63\%). But IWAL does better with a more principled approach, selecting homes that are likely to be most informative, with risk probabilities near 70\%. Figure \ref{fig:backtesting-100-even-hr-bytime-greedy-iwal-unif} shows how IWAL and two greedy heuristics differ. Higher HVI hit rate is not better; instead, it is the choice of which homes to explore with inspection that matters. The uncertainty in performace of each algorithm comes from sampling variation from running 25 independent simulated experiments. We prefer IWAL to alternatives because it has greater savings and is less sensitive to tuning parameters.

\emph{We acknowledge some assumptions in our simulations.} First, we only consider the cost of each job and not the time required for crews to move between homes, where there may be logistical issues with redirecting teams around the city. Second, in this analysis we have treated the \ActualFlint as having only 6,506 homes of which all are visited. This creates an arbitrary finite end point, as the algorithm runs out of homes with unsafe service lines. To avoid this effect, the above calculations, figures, and tables are based on the first 4,500 replacement visits and 2,250 hydrovac inspections. Of course, to validate this, we would need access to a larger set, and thus we turn to our larger simulation using a full size of Flint. Finally, the results are robust to resource allocation schedule and batch size. We recognize that we used a schedule of SLR and HVI activities different than Flint FAST Start. To disentangle the confound between our choice of algorithms and the schedule, we ran an additional version of the \ActualFlint backtest, with the schedule as closely aligned with Flint FAST Start in 2016-17 as possible. Across alternative scenarios tested the results differed only slightly.

\subsubsection{Results from \SimulatedFlint}

In our second simulation, we demonstrate the potential value of deploying the algorithm at scale and characterize the long-term performance of the algorithms. Via \SimulatedFlint we find that the proposed algorithms, with the aim of replacing hazardous lines from 18,000 homes out of a simulated city of 48,000 homes, can achieve 11.8\% savings relative to the current rate of spending. The best algorithm using IWAL yields an average effective cost of \$5,133 per successful replacement, better than  \$5,818 observed in Flint (Table \ref{tbl:costs}). As a final note, the proposed algorithms' SLR hit rates are all above 98.0\%. 

\singlespacing

\section*{Acknowledgments}

The authors would like to thank the FAST Start team for their phenomenal work and openness to collaboration. This includes Brigadier General (Ret.) Michael McDaniel, Ryan Doyle, Major Nicholas Anderson, and Kyle Baisden. Professors Lutgarde Raskin and Terese Olson, environmental engineering faculty at U-M, provided invaluable scientific support throughout. We are incredibly grateful to the work of, and communication with, Professor Martin Kaufman and Troy Rosencrantz at U-M Flint's GIS Center. We would like to thank Captricity, especially their machine learning team, Michael Zamora, Michael Zamora, David Shewfelt, and Kayla Pak for making the data accessible, and Kuang Chen for the generous support. We had major support from Mark Allison and his team of U-M Flint students. Rebecca Pettengill was enormously generous with her time and ability to help in the Flint community. We thank U-M Professors Marc Zimmerman and Rebecca Cunningham for their encouraging and helpful discussions. Among the many students involved in this work, we would like to recognize the roles of Jonathan Stroud and Chengyu Dai. And this work would not have happened without the expertise and enthusiasm of the students in the Michigan Data Science Team (MDST, \url{http://midas.umich.edu/mdst/}, \cite{farahi2018mdst}). The authors appreciate the many seminar and conference participants at U-M and elsewhere for their feedback on the academic work. The authors gratefully acknowledge the financial support of the Michigan Institute for Data Science (MIDAS), U-M's Ross School of Business, Google.org, and National Science Foundation CAREER grant IIS 1453304.


\bibliographystyle{ACM-Reference-Format}
\balance
\bibliography{flintrefs}

\clearpage

\end{document}